\documentclass[10pt,twocolumn,letterpaper]{article}

\usepackage{cvpr}              

\usepackage{graphicx}
\usepackage{amsmath}
\usepackage{amssymb}
\usepackage{booktabs}
\usepackage{multirow}

\usepackage[dvipsnames]{xcolor}
\usepackage{wrapfig,lipsum,booktabs}

\usepackage[pagebackref,breaklinks,colorlinks]{hyperref}
\usepackage{multicol}

\usepackage[capitalize]{cleveref}
\crefname{section}{Sec.}{Secs.}
\Crefname{section}{Section}{Sections}
\Crefname{table}{Table}{Tables}
\crefname{table}{Tab.}{Tabs.}

\renewcommand{\paragraph}[1]{\vspace{1mm}\noindent\textbf{#1}.}

\def\cvprPaperID{469} 
\def\confName{CVPR}
\def\confYear{2022}
\begin{document}

\title{MAD: A Scalable Dataset for Language Grounding in Videos\\from Movie Audio Descriptions }

{\author{
Mattia Soldan$^{1}$, 
\quad Alejandro Pardo$^{1}$,
\quad Juan León Alcázar$^{1}$,
\quad Fabian Caba Heilbron$^{2}$, \\
\quad Chen Zhao$^{1}$,
\quad Silvio Giancola$^{1}$, 
\quad Bernard Ghanem$^{1}$
\and
$^{1}$King Abdullah University of Science and Technology (KAUST) \quad $^{2}$Adobe Research\\
{\tt\small \{mattia.soldan, alejandro.pardo, juancarlo.alcazar, chen.zhao,}\\ 
{\tt\small silvio.giancola, bernard.ghanem\}@kaust.edu.sa}
\quad {\tt\small caba@adobe.com}
}}

\maketitle

\begin{abstract}
The recent and increasing interest in video-language research has driven the development of large-scale datasets that enable data-intensive machine learning techniques. In comparison, limited effort has been made at assessing the fitness of these datasets for the video-language grounding task. Recent works have begun to discover significant limitations in these datasets, suggesting that state-of-the-art techniques commonly overfit to hidden dataset biases. In this work, we present MAD (Movie Audio Descriptions), a novel benchmark that departs from the paradigm of augmenting existing video datasets with text annotations and focuses on crawling and aligning available audio descriptions of mainstream movies. MAD contains over $384,000$ natural language sentences grounded in over $1,200$ hours of videos and exhibits a significant reduction in the currently diagnosed biases for video-language grounding datasets. 
MAD's collection strategy enables a novel and more challenging version of video-language grounding, where short temporal moments (typically seconds long) must be accurately grounded in diverse long-form videos that can last up to three hours.
We have released MAD's data and baselines code at \url{https://github.com/Soldelli/MAD}.

\vspace{-.5cm}
\end{abstract}
\section{Introduction}\label{sec: intro}
\begin{figure}[th]
    \centering
        \includegraphics[trim={0.1cm 0.2cm 0.2cm 0.2cm},width=\linewidth,clip]{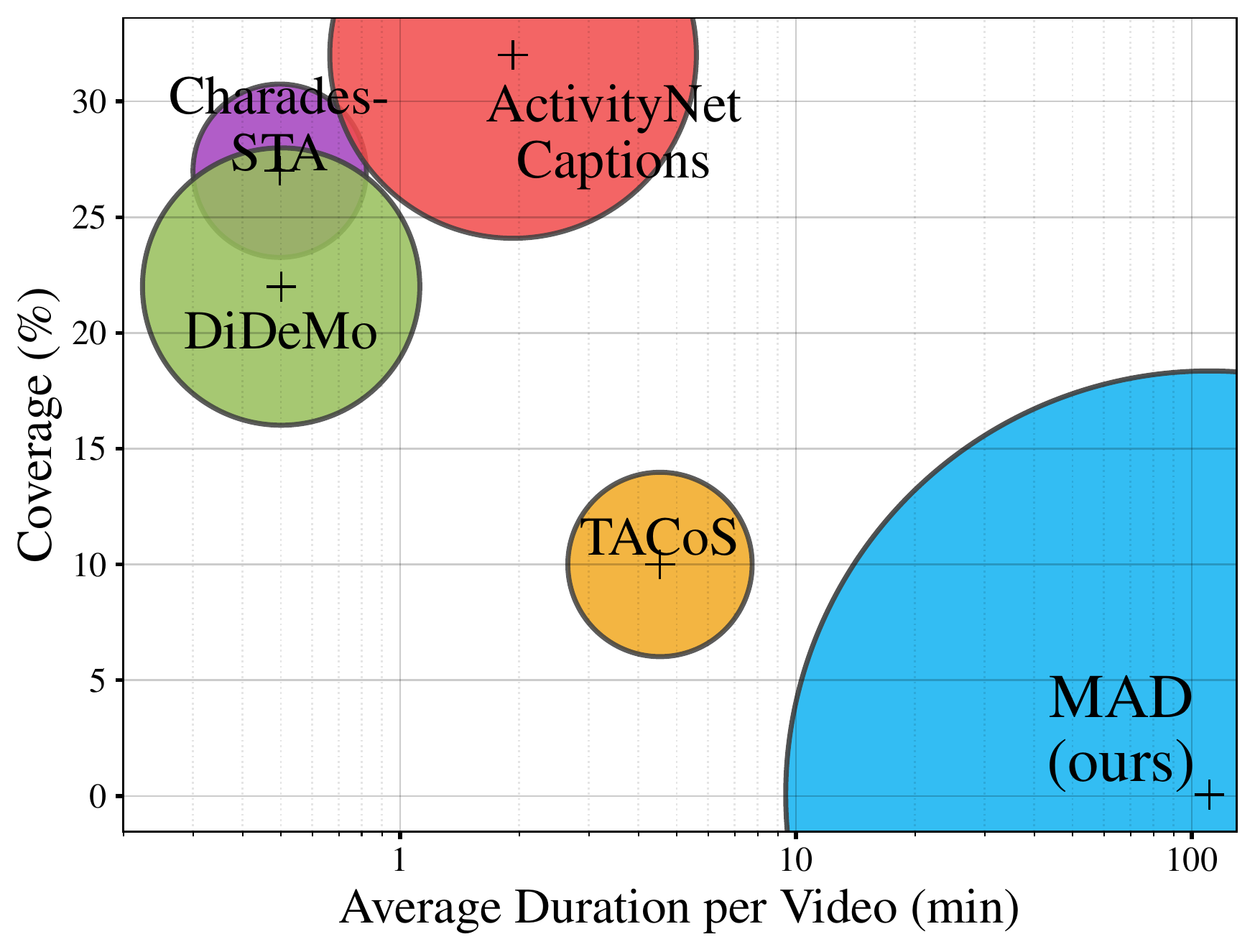}
        \caption{
        \textbf{Comparison of video-language grounding datasets.} 
        The circle size measures the language vocabulary diversity.
        The videos in MAD are orders of magnitude longer in duration than previous datasets (${\sim}110$min), annotated with natural, highly descriptive, language grounding (${>}60$K unique words) with very low coverage in video (${\sim}4.1$s). Coverage is defined as the average \% duration of moments with respect to the total video duration.
    }
    \label{fig:PullingFigure}
    \vspace{-.5cm}
\end{figure}

Imagine you want to find the moment in time, in a movie, when your favorite actress is eating a Gumbo dish in a New Orleans restaurant. You could do so by manually scrubbing the film to ground the moment. However, such a process is tedious and labor-intensive. This task is known as natural language grounding~\cite{Hendricks_2017_ICCV,Gao_2017_ICCV}, and has gained significant momentum in the computer vision community. Beyond smart browsing of movies, the interest in this task stems from multiple real-world applications ranging from smart video search~\cite{sivic2003video, snoek2009mediamill}, video editing~\cite{xiong2021transcript,pardo2021learning,pardo2021moviecuts}, and helping patients with memory dysfunction~\cite{budson2005memory, toyama2015towards}. The importance of solving this task has resulted in novel approaches and large-scale deep-learning architectures that steadily push state-of-the-art performance.

\begin{figure*}[!ht]
    \vspace{-0.2cm}
    \center
    \includegraphics[trim={0.cm 0.cm 0.cm 0cm},width=0.9\linewidth,clip]{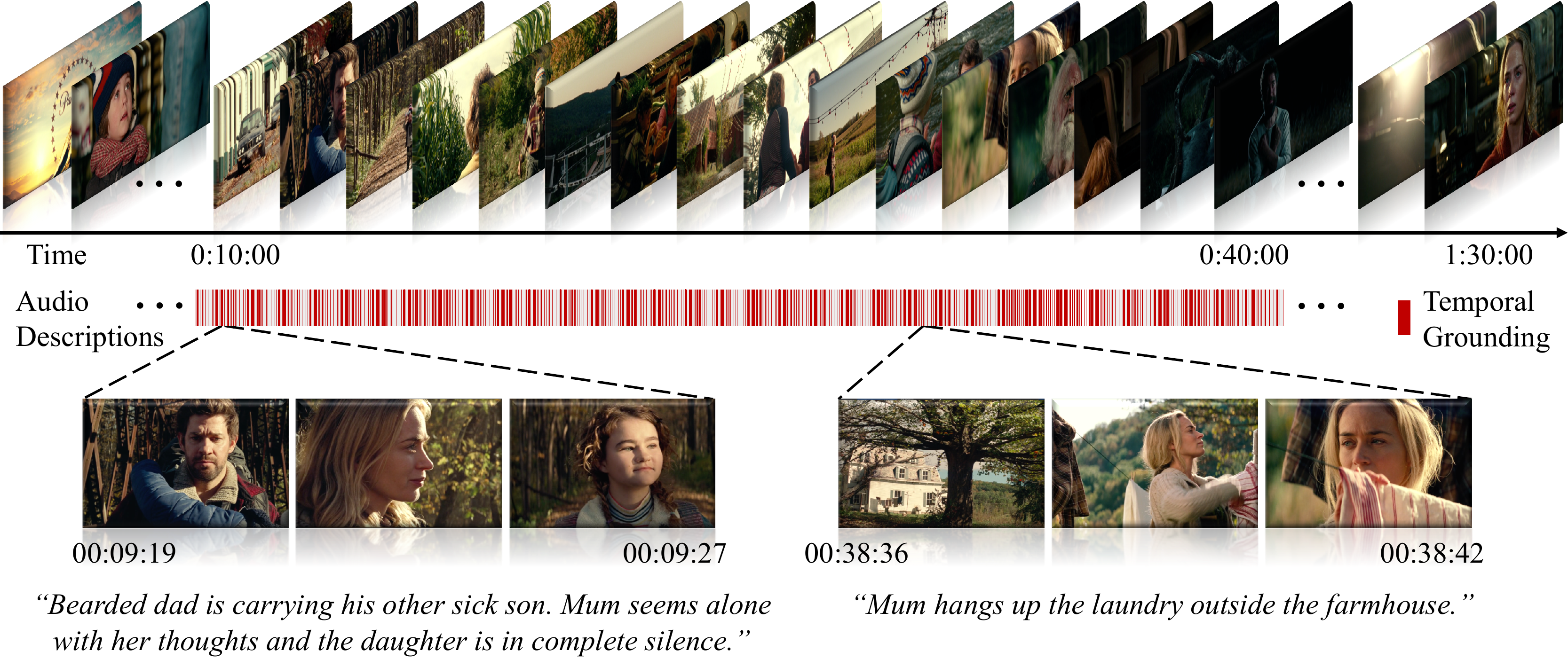}
    \caption{\textbf{Example from our MAD dataset.} 
    We select the movie ``A quiet place'' as representative for our dataset. As shown in the figure, the movie contains a large number of densely distributed temporally grounded sentences.
    The collected annotations can be very descriptive, mentioning people, actions, locations, and other additional information. Note that as per the movies plot, the characters are silent for the vast majority of the movie, rendering audio description essential for visually-impaired audience.
    }
    \label{fig:mad-sample}
    \vspace{-0.4cm}
\end{figure*}

Despite those advances, recent works~\cite{otani2020challengesmr, DBLP:journals/corr/abs-2101-09028,zhang2021towards} have diagnosed hidden biases in the most common video-language grounding datasets. 
Otani~\etal~\cite{otani2020challengesmr} have highlighted that several grounding methods only learn location priors, to the point of disregarding the visual information and only using language cues for predictions. 
While recent methods~\cite{DBLP:journals/corr/abs-2101-09028,zhang2021towards} have tried to circumvent these limitations by either proposing new metrics~\cite{yuan2021closer} or debiasing strategies~\cite{zhang2021towards,zhou2021embracing}, it is still unclear if existing grounding datasets~\cite{TACoS_ACL_2013,Krishna_2017_ICCV,Hendricks_2017_ICCV,Gao_2017_ICCV} provide the right setup to evaluate progress in this important task.
This is partly because datasets used by video-language grounding models were not originally collected to solve this task.
These datasets provide valuable video-language pairs for captioning or retrieval, but the grounding task requires high-quality (and dense) temporal localization of the language.
Figure~\ref{fig:PullingFigure} shows that the current datasets comprise relatively short videos, containing single structured scenes, and language descriptions that cover most of the video. 
Furthermore, the temporal anchors for the language are temporally biased, leading to methods not learning from any visual features and eventually overfitting to temporal priors for specific actions, thus limiting their generalization capabilities~\cite{otani2020challengesmr,lan2021survey}.

In this work, we address these limitations with a novel large-scale dataset called MAD (Movie Audio Descriptions). MAD builds atop (and includes part of) the LSMDC dataset~\cite{rohrbach2017movie}, which is a pioneering work in leveraging audio descriptions to enable the investigation of a closely related task: text-to-video retrieval. 
Similar to LSMDC, we depart from the standard annotation pipelines that rely on crowd-sourced annotation platforms. Instead, we adopt a scalable data collection strategy that leverages professional, grounded audio descriptions of movies for visually impaired audiences. As shown in Figure~\ref{fig:mad-sample}, MAD grounds these audio descriptions on long-form videos, bringing novel challenges to the video-language grounding task.

Our data collection approach consists of transcribing the audio description track of a movie and automatically detecting and removing sentences associated with the actor's speech, yielding an authentic ``untrimmed video'' setup where the highly descriptive sentences are grounded in long-form videos.
Figures~\ref{fig:PullingFigure} and~\ref{fig:dataset-attributes} illustrate the uniqueness of our dataset with respect to the current alternatives.  
As showcased in Figure~\ref{fig:mad-sample}, MAD contains videos that, on average, span over $110$ minutes, as well as grounded annotations covering short time segments, which are uniformly distributed in the video, and maintain the largest diversity in vocabulary.
Video grounding in MAD requires a finer understanding of the video, since the average coverage of the sentences is much smaller than in current datasets. 

The unique configuration of the MAD dataset introduces exciting challenges. 
First, the video grounding task is now mapped into the domain of long form video, preventing current methods to learn trivial temporal location priors, instead requiring a more nuanced understanding of the video and language modalities.
Second, having longer videos means producing a larger number of segment proposals, which will make the localization problem far more challenging. 
Last, these longer sequences emphasize the necessity for efficient methods in inference and training, mandatory for real-world applications such as long live streaming videos or moment retrieval in large video collections~\cite{escorcia2019temporal}. 

\paragraph{Contributions} Our contributions are threefold.
\textbf{(1)} We propose Movie Audio Description (MAD), a novel large-scale dataset for video-language grounding, containing more than $384$K natural language sentences anchored on more than $1.2$K hours of video. 
\textbf{(2)} We design a scalable data collection pipeline that automatically extracts highly valuable video-language grounding annotations, leveraging speech-to-text translation on professionally generated audio descriptions.
\textbf{(3)} We provide a comprehensive empirical study that highlights the benefits of our large-scale MAD dataset on video-language grounding as a benchmark, pointing out the difficulties faced by current video-language grounding baselines in long-form videos. 

\section{Related work}\label{sec: related}

\paragraph{Video Grounding Benchmarks}
Most of the current video grounding datasets were previously collected and tailored for other computer vision tasks (\ie, Temporal Activity Localization~\cite{7298698, TACoS_ACL_2013}) and purposes (\ie, Human Action Recognition), then annotated for the video grounding task. This adaptation limits the diversity of the video corpus towards a specific set of actions and objects, and its corresponding natural language sentences to specific sets of verbs and nouns~\cite{rohrbach2014coherent, Krishna_2017_ICCV, Gao_2017_ICCV, otani2020challengesmr}. Currently, ActivityNet-Captions~\cite{Krishna_2017_ICCV} and Charades-STA~\cite{Gao_2017_ICCV} are the most commonly used benchmarks for the task of video grounding. Both datasets have been collected atop pre-existing video datasets (ActivityNet~\cite{7298698} and Charades~\cite{10.1007/978-3-319-46448-0_31}) and have been diagnosed with severe biases by Otani~\etal~\cite{otani2020challengesmr}. These findings show that the annotations contain distinct biases where language tokens are often coupled with specific temporal locations. Moreover, strong priors also affect the temporal endpoints, with a large portion of the annotations spanning the entire video. As a consequence, current methods seem to mainly rely on such biases to make predictions, often disregarding the visual input altogether. In comparison, the unique setup of MAD prevents these drawbacks as keywords are not associated with particular temporal regions, and annotation timestamps are much shorter than the video's duration.  Moreover, different from Charades-STA, MAD defines an official validation set for hyper-parameter tuning. 

Unlike Charades-STA and ActivityNet-Captions, TACoS \cite{TACoS_ACL_2013} has not been diagnosed with annotation biases. However, its video corpus is small and limited to cooking actions recorded in a static-camera setting. Conversely, spanning over $22$ genres across $90$ years of cinema history, MAD covers a broad domain of actions, locations, and scenes. 
Moreover, MAD inherits a diverse set of visual and linguistic content from the broad movie genres, ranging from fiction to everyday life.

Furthermore, DiDeMo~\cite{Hendricks_2017_ICCV} was annotated atop Flickr videos with a discrete annotation scheme (\ie in chunks of $5$ seconds) for a maximum of $30$ seconds, constraining the problem of video grounding to trimmed videos. Given these annotations, the grounding task can be simplified to choosing one out of $21$ possible proposals for each video. 
Conversely, MAD provides a setup to explore solutions for grounding language in long-form videos, whose length can be up to $3$ hours.  
In this scenario, naive sliding window techniques for proposal generation could produce  hundreds of thousands of possible  candidates. Therefore, developing efficient inference methods becomes a much more urgent requirement compared to previous benchmarks.

State-of-the-art video grounding methods~\cite{2DTAN_2020_AAAI, soldan2021vlg, liu2020jointly, Zeng_2020_CVPR, Mun_2020_CVPR,Zhao_2021_CVPR,Liu_2021_CVPR} have relied on existing benchmarks to design novel modules (\ie, proposal generation, context modeling, and multi-modality fusion). However, most of these designs specifically target grounding in short videos and often rely on providing the entire video to the model when making a prediction. As the long-form setup introduced by MAD prohibits this, new methods will have the opportunity to investigate and bridge previous ideas to these new challenging and real-world  constraints.

\noindent\textbf{Audio Descriptions}.
The pioneering works of Rohrbach \etal~\cite{rohrbach2015dataset} and Torabi \etal~\cite{torabi2015using} were the first to exploit audio descriptions to study the text-to-video retrieval task and its counterpart video-to-text. 
Rohrbach~\etal~\cite{rohrbach2015dataset} introduced the MPII-MD dataset, while Torabi~\etal~\cite{torabi2015using} presented M-VAD, both collected from audio descriptions in movies. Later, these datasets were fused to create the LSMDC dataset~\cite{rohrbach2017movie}, which forms the core of the LSMDC annual challenge. Our annotation pipeline is similar in spirit to these works. However, MAD sizes the potential of movies' audio descriptions for a new scenario: language grounding in videos.

A concurrent work introduced a new benchmark based on audio descriptions in videos, called QuerYD~\cite{9414640}. It is a dataset for retrieval and event localization in videos crawled from YouTube. This benchmark focuses on the short-form video setup with videos of less than $5$ minutes average duration. QuerYD also leverages audio descriptions, which have been outsourced to volunteer narrators. Similar to our takeaways, the authors noticed that audio descriptions are generally more visually grounded and descriptive than previously collected annotations.

\section{Collecting the MAD Dataset}\label{sec: method}

In this section, we outline MAD's data collection pipeline. We follow independent strategies for creating the training and testing set. For the former, we aim at automatically collecting a large set of annotations. For the latter, we re-purpose the manually refined annotations in LSMDC. We provide detailed statistics of MAD's annotations and compare them to existing  datasets.

\subsection{MAD Training set}
\label{section::train-set}

\begin{table*}[ht]
    \centering
    \resizebox{\linewidth}{!}{
    \begin{tabular}{l||r|r|r||r|r|r|r|r|r|r}
                                      & \multicolumn{3}{c||}{Videos}    & \multicolumn{7}{c}{Language Queries}       \\ \cmidrule{2-11}
Dataset                                &  Total    & Duration   & Duration   &  Total   & \# Words  &  Total   & \multicolumn{4}{c}{Vocabulary}\\ \cline{8-11} 
                                       & Duration  & / Video  & / Moment &  Queries & / Query &  Tokens  & Adj. & Nouns & Verbs & Total  \\ \midrule
TACoS~\cite{TACoS_ACL_2013}            &    $10.1$ h &   $4.78$ min & $27.9$  s    &   $18.2$K  & $10.5$      &  $0.2$M    & $0.2$K &  $0.9$K &  $0.6$K &  $2.3$K  \\
Charades-STA~\cite{Gao_2017_ICCV}      &    $57.1$ h &   $0.50$ min &  $8.1$  s    &   $16.1$K  & $7.2$       &  $0.1$M    & $0.1$K &  $0.6$K &  $0.4$K &  $1.3$K  \\
DiDeMo~\cite{Hendricks_2017_ICCV}      &    $88.7$ h &   $0.50$ min &  $6.5$  s    &   $41.2$K  & $8.0$       &  $0.3$M    & $0.6$K &  $4.1$K &  $1.9$K &  $7.5$K  \\
ANet-Captions~\cite{Krishna_2017_ICCV} &   $487.6$ h &   $1.96$ min & $37.1$  s    &   $72.0$K  & $14.8$      &  $1.0$M    & $1.1$K &  $7.4$K &  $3.7$K & $15.4$K  \\ \midrule
\textbf{MAD (Ours)}                    &  $1207.3$ h & $110.77$ min &  $4.1$  s    &  $384.6$K  & $12.7$      &  $4.9$M    & $5.3$K & $35.5$K & $13.1$K & $61.4$K  \\
\bottomrule
    \end{tabular}
    }
    \caption{
    \textbf{Statistics of video-language grounding datasets.}
    We report relevant statistics to compare our MAD dataset against other video grounding benchmarks.  
    MAD provides the largest dataset with $1207$hrs of video and $384.6$K language queries, the longest form of video (avg. $110.77$min), the most diverse language vocabulary with $61.4$K unique words, and the shortest moment for grounding (avg. $4.1$s).
    }
    \label{tab:datasets}
    \vspace{-0.2cm}
\end{table*}

MAD relies on audio descriptions professionally created to make movies accessible to visually-impaired audiences. These descriptions embody a rich narrative describing the most relevant visual information. Thus, they adopt a highly descriptive and diverse language. Audio descriptions are often available as an alternative audio track that can replace the original one. Professional narrators curate them, and devote significant effort to describe a movie. The audio description process demands an average of $30$ work hours to narrate a single hour of video~\cite{rohrbach2017movie}. In comparison, previous datasets that have used the Amazon Mechanical Turk service for video-language grounding estimate the annotation effort to be around $3$ hours for each video hour~\cite{Krishna_2017_ICCV}.

\paragraph{Data Crawling} Not every commercially available movie is released with audio descriptions. However, we can obtain these audio descriptions from 3$^{\text{rd}}$ party creators. In particular, we crawl our audio descriptions from a large open-source and online repository\footnote{\url{https://www.audiovault.net/}}. 
These audio files contain the original movie track mixed with the narrator's voice, carefully placed when actors are not speaking.
One potential problem is that the audio descriptions can be misaligned with the original movie. Such misalignment comes either from a delay in the recording of the audio description (concerning the original movie) or from audio descriptions being created from different versions of the movie (with deleted or trimmed scenes). 

\paragraph{Alignment and Cleanup} Since the audio description track also contains the movie's original audio, we can resolve this misalignment by maximizing the cross-correlation between overlapping segments of the original audio track and the audio description track. We define the original audio signal $(f)$, the audio description signal $(g)$, and the time delay $(\tau_{\mathrm{delay}})$ between the two signals. The maximum of the cross-correlation function (denoted with the operator $\star$) indicates the point in time where the signals exhibit the best alignment. As a result, the time delay $\tau_{\mathrm{delay}}$ between $f$ and $g$ is defined as follows: 

\vspace{-0.2cm}
\begin{equation}
    \tau _{\mathrm {delay} }={\underset {t}{\arg\max}~~((f \star g)(t))}    \label{eq:Xcorr}
\end{equation}

\begin{figure*}[ht!]
    \vspace{-0.2cm}
    \centering
    \begin{subfigure}[t]{0.325\linewidth}
        \centering
        \includegraphics[width=\linewidth]{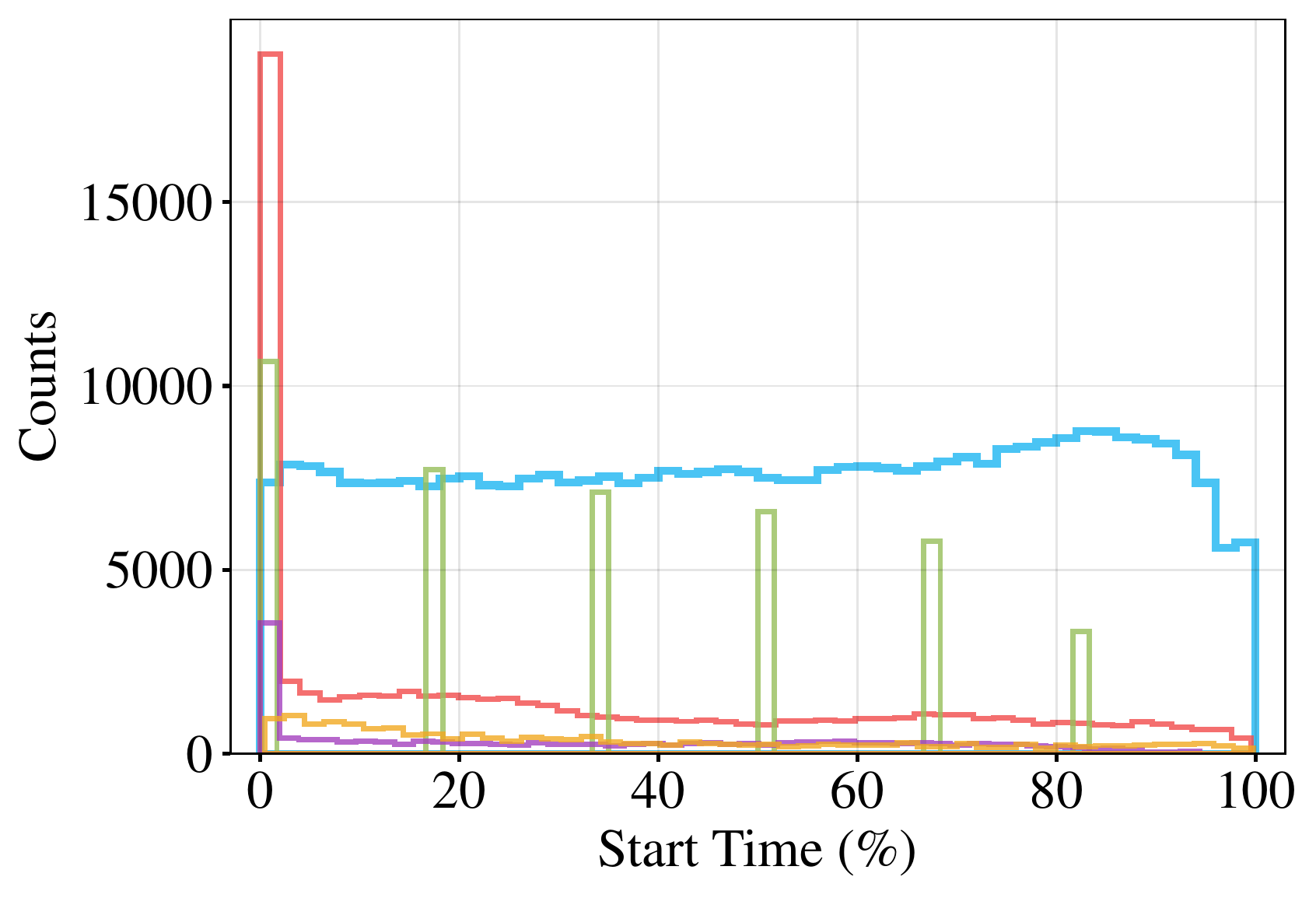}
        \caption{Moment start time histogram}
        \label{fig:starts}
    \end{subfigure}%
    \begin{subfigure}[t]{0.31\linewidth}
        \centering
        \includegraphics[width=\linewidth]{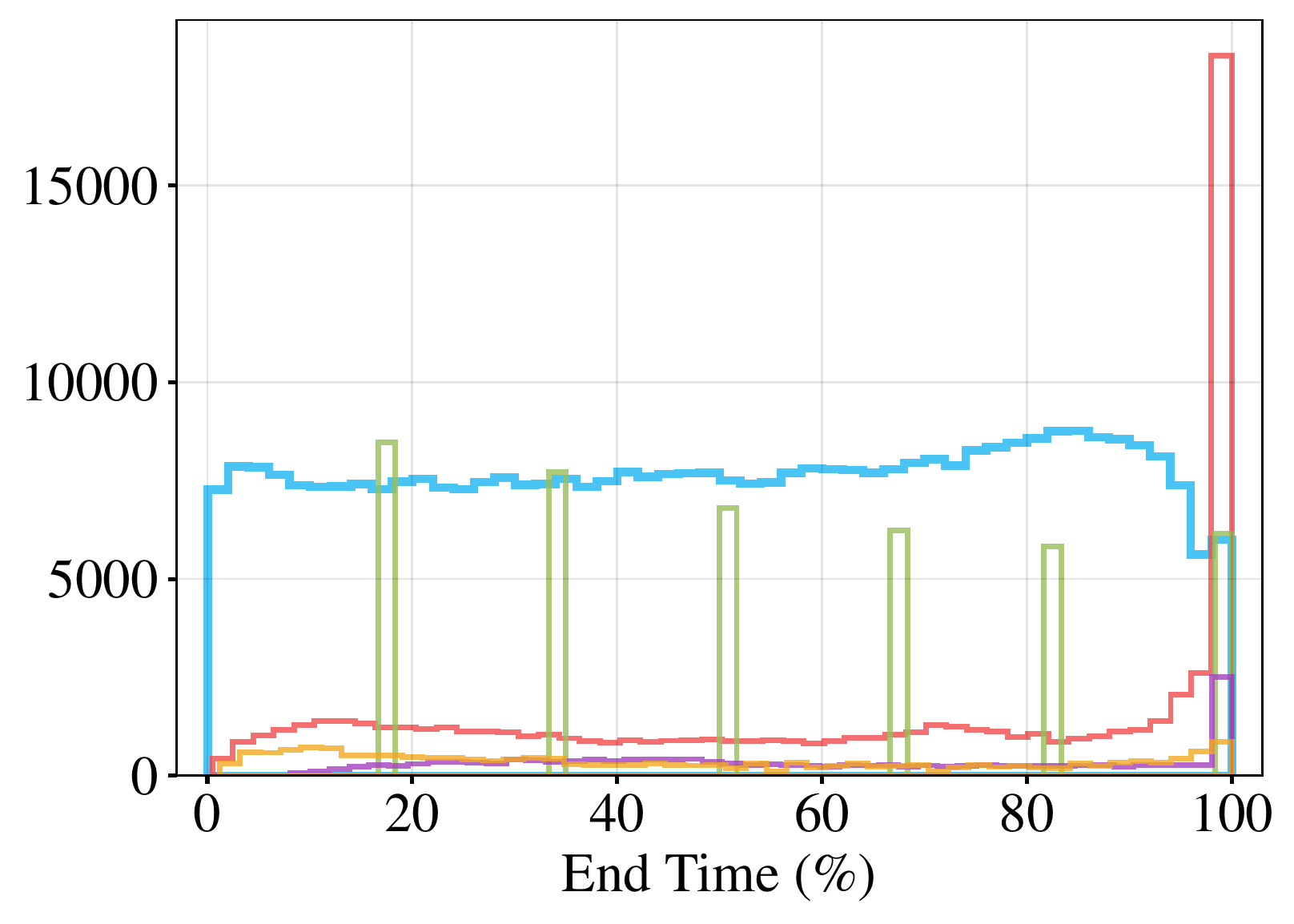}
        \caption{Moment end time histogram}
        \label{fig:ends}
    \end{subfigure}%
    \begin{subfigure}[t]{0.35\linewidth}
        \centering
        \includegraphics[trim=0 0 2.3cm 0.9cm, width=\linewidth]{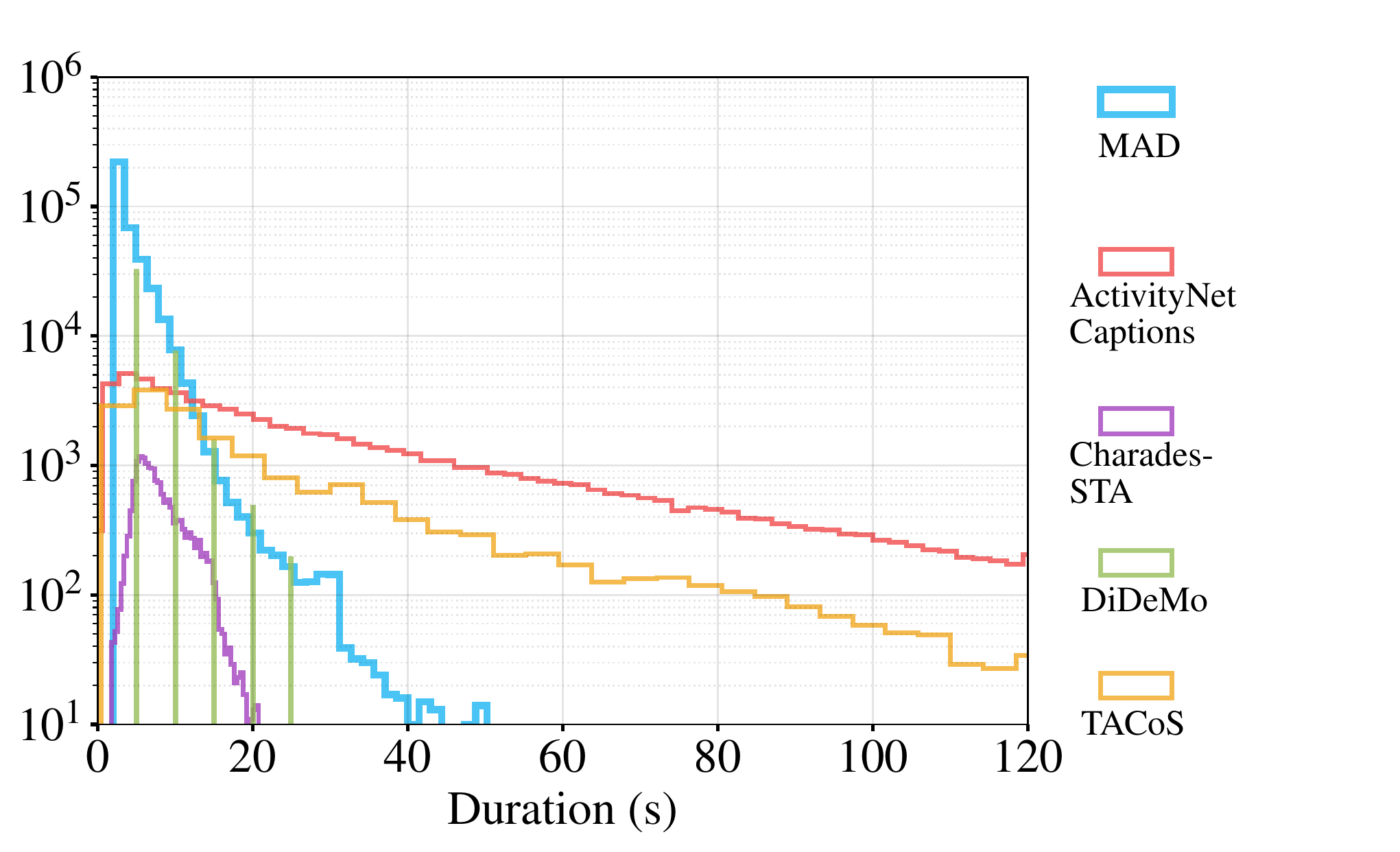}
        \caption{Moment duration histogram}
        \label{fig:durations}
    \end{subfigure}
    \caption{\textbf{Histograms of moment start/end/duration in video-language grounding datasets.} The plots represent the normalized (by video length) start/end histogram (a-b) and absolute duration distribution (c) for moments belonging to each of the five datasets. We notice severe biases in ActivityNet-Captions and Charades-STA, which show high peaks at the beginning and end of the videos. Conversely MAD does not show any particular preferred start/end temporal location.
    }
    \label{fig:dataset-attributes}
    \vspace{-0.4cm}
\end{figure*}

To verify that our single time delay $\tau_{\mathrm{delay}}$ defines the best possible alignment between the audio descriptions and the original movies, we run our synchronization strategy over several (i.e., $20$) uniformly distributed temporal windows. 
We verify that the delays estimated for all windows are consistent with each other, within a maximum range of $\pm0.1$ seconds w.r.t the median value of the distribution of the $20$ samples. We discard the movies that do not satisfy this criterion to ensure that the whole audio description track correctly aligns with the original movie's visual content.

\paragraph{Audio Transcriptions and Verification}
After the two audio tracks are aligned, we transcribe the audio description file using Microsoft's Azure Speech-to-Text service\footnote{\url{https://azure.microsoft.com/en-us/services/cognitive-services/speech-to-text/}}. Each recognized word is associated with a temporal timestamp. At this step in our pipeline, we have sentences temporally grounded to the original video stream. As the Speech-to-Text contains both the narrations and the actors' speech, we set out to remove the latter and only retain the audio description in textual form.
To do so, we resort to the movie's original subtitles and use their timestamps as a surrogate for Voice Activity Detection (VAD). Particularly, we discard closed captions and retain subtitles associated with the actors' speech or subtitles for songs. Then, we remove from the Speech-to-Text output every sentence overlapping with the VAD temporal locations, obtaining our target audio description sentences. We post-process the text for automatic punctuation refinement using the free tool Punctuator~\cite{tilk2016}.

\subsection{From LSMDC to MAD Val/Test}
\label{section::test-set}
As outlined in Section~\ref{section::train-set}, the val/test sets of MAD also relies on audio descriptions for movies. Since the annotations in training  are automatically generated, we decided to minimize the noise in the validation and test splits. Hence, we avoid the automatic collection of data for these sets and resort to the data made available by the LSMDC dataset~\cite{rohrbach2017movie}. This dataset collected annotations from audio descriptions in movies targeting the video retrieval task. LSMDC manually refined the grammar and temporal boundaries of sentences. As a consequence, these annotations have clean language and precise temporal boundaries. We reformat a subset of the LSMDC data, adapt it for the video grounding task, and cast it as MAD's val and test sets.

LSMDC data for retrieval is made available only as video chunks, not full movies. To create data suitable for long-form video language grounding, we collect $162$ out of the $182$ videos in LSMDC and their respective audio descriptions. Again, the full-length video data and the chunked video data provided by LSMDC might not be in sync. To align a video chunk from LSMDC with our full-length movies, we follow a similar procedure as the one described in Section \ref{section::train-set} for the audio alignment, but using visual information. We use CLIP \cite{radford2021learning} to extract five frame-level features per second for both the video chunks from LSMDC and our full-length movie. Then, we use the maximum cross-correlation score to estimate the delay between the two. We run the alignment on $10$ different window lengths and take the median value as the delay of the chunk. 

Once the audiovisual data is aligned, we use the text annotations and re-define their original timestamps according to the calculated delays. This process creates full-length movies with curated grounding data for MAD's validation and test sets. In doing so, we obtain large and clean validation and test sets since LSMDC was curated by humans and has been used for years by the community. Our val/test sets evaluate video-grounding methods with more than $104$K grounded phrases coming from more than $160$ movies. 

\begin{table*}[!ht]
    \centering
\setlength{\tabcolsep}{2.5pt}
\renewcommand{\arraystretch}{1} 
\resizebox{\linewidth}{!}{%
\footnotesize
\begin{tabular}{@{}c@{\hspace{0.4em}} l 
c@{\hspace{0.2em}} 
cccc c@{\hspace{0.2em}} 
ccccc c@{\hspace{0.2em}} 
ccccc c@{\hspace{0.2em}} 
c@{}}
\toprule
&&\phantom{} & \multicolumn{5}{c}{IoU=0.1} 
& \phantom{} & \multicolumn{5}{c}{IoU=0.3} 
& \phantom{} & \multicolumn{5}{c}{IoU=0.5} 
& \phantom{} \\
\cmidrule{4-8} \cmidrule{10-14} \cmidrule{16-21} 
& Model   && R@1 & R@5 & R@10 & R@50 & R@100
          && R@1 & R@5 & R@10 & R@50 & R@100 
          && R@1 & R@5 & R@10 & R@50 & R@100 \\
\midrule
&Oracle   &&$100.00$ &  $-$  &  $-$  &  $-$  &  $-$   
          &&$100.00$ &  $-$  &  $-$  &  $-$  &  $-$  
          &&$99.99$  &  $-$  &  $-$  &  $-$  &  $-$ &\\
          
&Random Chance   && $0.09$  &  $0.44$ & $0.88$ & $4.33$ & $8.47$ 
          && $0.04$  &  $0.19$ & $0.39$ & $1.92$ & $3.80$ 
          && $0.01$  &  $0.07$ & $0.14$ & $0.71$ & $1.40$  \\
          
&CLIP~\cite{radford2021learning}  && $\mathbf{6.57}$ & $\mathbf{15.05}$ & $\mathbf{20.26}$ & $37.92$ & $47.73$ 
                                  && $\mathbf{3.13}$ & $\mathbf{9.85}$  & $14.13$ & $28.71$ & $36.98$ 
                                  && $1.39$ &  $5.44$ &  $8.38$ & $18.80$ & $24.99$  \\
                                  
&VLG-Net~\cite{soldan2021vlg}     && $3.64$           & $11.66$ & $17.89$ & $\mathbf{39.78}$ & $\mathbf{51.24}$ 
                                  && $2.76$ & $9.31$ & $\mathbf{14.65}$ & $\mathbf{34.27}$ & $\mathbf{44.87}$ 
                                  && $\mathbf{1.65}$ & $\mathbf{5.99}$ & $\mathbf{9.77}$ & $\mathbf{24.93}$ & $\mathbf{33.95}$ \\
\bottomrule
\end{tabular}
}
\vspace{-.1cm}
\caption{\label{tab:mad_grounding}{\bf Benchmarking of grounding baselines on the MAD dataset.}  
We report the performance of four baselines: \textit{Oracle}, \textit{Random  Chance}, \textit{CLIP}, \textit{VLG-Net}, on the test split. The first two validate the choice of proposals by computing the upper bound to the performance and the random performance. CLIP and VLG-Net use visual and language features to score and rank proposals. For all experiments, we adopt the same proposal scheme as in VLG-Net~\cite{soldan2021vlg}, and use CLIP~\cite{radford2021learning} features for video (frames) and language embeddings. 
}
\vspace{-.3cm}

\end{table*}

\subsection{MAD Dataset Analysis} 
We now present MAD's statistics and compare them with legacy datasets for the task. Table~\ref{tab:datasets} summarizes the most notable aspects of MAD regarding both the visual and language content of the dataset. 

\paragraph{Scale and Scalability} MAD is the largest dataset in video hours and number of sentences. The training, validation, and test sets consist of $488$, $50$, and $112$ movies, respectively.  Although other datasets have a larger number of clips, MAD's videos are full movies which last $2$ hours on average. In comparison, the average clip from other datasets spans just a few minutes. Overall, MAD splits contain a combined $50$ days of continuous video. We highlight that our test set alone is already larger than any other video grounding dataset. We also emphasize that each movie in MAD is long and composed of several diverse scenes, making it a rich source for long-form video analysis. Also, as the cinema industry is ever-growing, we expect to periodically expand MAD with subsequent releases to fuel further innovation in this research direction. 

\paragraph{Vocabulary Size} Besides having the largest video corpus, MAD also contains the largest and most diverse query set of any dataset. In Table~\ref{tab:datasets}, we show that MAD contains the largest set of adjectives, nouns, and verbs among all available benchmarks. In almost every aspect, it is an order of magnitude larger. The number of sentences in MAD training, validation, and test is $280.5$K, $32.1$K, and $72.0$K, respectively, one order of magnitude larger than the equivalent set in any other dataset. Overall, MAD contains $61.4$K unique words, almost $4$ times more than the $15.4$K of ActivityNet-Captions~\cite{Krishna_2017_ICCV} (the highest among the other benchmarks). Finally, the average length per sentence is $12.7$ words, which is similar to the other datasets.

\paragraph{Bias Analysis} Figure~\ref{fig:dataset-attributes} plots the histograms for start/end timestamps of moments in all grounding datasets. We notice clear biases in current datasets: Charades-STA~\cite{Gao_2017_ICCV}, DiDeMo~\cite{Hendricks_2017_ICCV}, and ActivityNet-Captions~\cite{Krishna_2017_ICCV}. 
Charades-STA and ActivityNet-Captions are characterized by tall peaks at the beginning (Figure~\ref{fig:starts}) and end (Figure~\ref{fig:ends}) of the video, meaning that most temporal annotations start at the video's start or finish at the video's end. This bias is learned quickly by modern machine learning algorithms, resulting in trivial groundings that span a full video. The smaller dataset TACoS also exhibits a similar bias, although it is less prominent. DiDeMo is limited by its annotation strategy, where chunks of $5$ seconds are labeled up to a maximum of $30$ seconds. This favors structured responses that roughly approximate the start and end points of a moment. In contrast, MAD has an almost uniform histogram. This means that moments of interest can start and end at any point in the video. 
We only observe a minor imbalance where the end of the movie has slightly more descriptions than the beginning. This is related to the standard structure of a film, where the main plot elements are resolved towards the end, thus creating more situations worth describing. 
Figure~\ref{fig:durations} plots the histograms for moment duration. MAD is characterized by shorter moments on average, having a long tail distribution with moments that last up to one minute.

\section{Experiments}\label{sec: experiments}
We proceed with the experimental assessment of video-language grounding on the MAD dataset. We first describe the video grounding task in the MAD dataset along with its evaluation metrics and then report the performance of four selected baselines.

\paragraph{Task}
Given an untrimmed video and a language query, the video-language grounding task aims to localize a temporal moment $(\tau_s, \tau_e)$ in the video that matches the query~\cite{Gao_2017_ICCV,Hendricks_2017_ICCV}. 

\paragraph{Metric}
Following the grounding literature~\cite{Gao_2017_ICCV,Hendricks_2017_ICCV}, we adopt Recall@$K$ for IoU${=}\theta$ (R@$K$-IoU${=}\theta$). Given a ranked set of video proposals, this metric measures if any of the top $K$ ranked moments have an IoU larger than $\theta$ with the ground truth temporal endpoints. Results are averaged across all test samples. Given the long-form nature of our videos and the large amount of possible proposals, we investigate Recall@$K$ for IoU${=}\theta$ with $K{\in}\{1,5,10,50,100\}$ and $\theta{\in}\{0.1, 0.3, 0.5\}$. 
This allows us to evaluate for loose alignment (\ie IoU${=}0.1$) and approximate ranking (\ie, $K{=}100$), as well as tight predictions (\ie IoU${=}0.5$) and accurate retrieval (\ie $K{=}1$). 

\paragraph{Baselines}
We benchmark MAD using four different grounding strategies, namely: \textit{Oracle}, \textit{Random Chance}, \textit{CLIP}~\cite{radford2021learning}, and \textit{VLG-Net}~\cite{soldan2021vlg}. The first two provide upper bounds and random performance for the recall metric given a predefined set of proposals. \textit{Oracle} chooses the proposal with the highest IoU with the ground-truth annotation, while \textit{Random Chance} chooses a random proposal with uniform probability. We also use \textit{CLIP}, the pre-trained image-text architecture from~\cite{radford2021learning}, to extract frame-level and sentence-level features. The frame-level features for each proposal are combined using mean pooling, then we score each proposal using cosine similarity between the visual and the text features. Finally, we adopt VLG-Net~\cite{soldan2021vlg} as a representative, state-of-the-art method for the  grounding task. VLG-Net leverages recent advances in Graph Convolution Networks (GCNs)~\cite{li2019deepgcns,xu2020gtad,li2021deepgcns_pami,li2020deepergcn,zhao2021video} to model individual modalities, while also enabling the aggregation of non-local and cross-modal context through graph convolutions. We use VLG-Net and adapt it to work with MAD's long-form videos. See the \textbf{supplementary material} for details. 

\paragraph{Implementation Details}
For the lower bound estimation using \textit{Random Chance}, we average the performance over $100$ independent runs. 
To favor a fair comparison against the CLIP baseline, we train VLG-Net using CLIP features for both modalities. We use the official VLG-Net's implementation with a clip size of $128$ input frames spanning $25.6$ seconds (frames are extracted at $5$ fps). We train using the Adam optimizer~\cite{kingma2014adam} with learning rate of $10^{-4}$. For inference over an entire movie, we adopt a sliding window approach, where we stride the input window by $64$ frames and discard highly redundant proposals through Non Maximum Suppression (NMS) with a threshold of $0.3$.

\subsection{Grounding Performance on MAD}
Table~\ref{tab:mad_grounding} summarizes the baseline performance on MAD. The  \textit{Oracle} evaluation achieves a perfect score across all metrics except for IoU${=}0.5$. Only a negligible portion of the annotated moments cannot be correctly retrieved at a high IoU ($0.5$), this result showcases the suitability of the proposal scheme. The low performance of the \textit{Random Chance} baseline  reflects the difficulty of the task, given the vast pool of proposals extracted over a single video. For the least strict metric (R@$100$-IoU${=}0.1$), this baseline only achieves  $8.47\%$, while CLIP and VLG-Net baselines are close to $50\%$, a ${\sim}6\times$ relative improvement. An even larger gap is present for the most strict metric, R@$1$-IoU${=}0.5$, with a relative improvement of two orders of magnitude.

The CLIP~\cite{radford2021learning} baseline is pre-trained for the task of text-to-image retrieval, and we do not fine-tune this model on the MAD dataset. Nevertheless, when evaluated in a zero-shot fashion, it results in a strong baseline for long-form grounding, achieving the best R@$K$ for the least strict IoU${=}0.1$ at $K{=}\{1,5,10\}$. Although IoU${=}0.1$ corresponds to very loose grounding, this result is nonetheless valuable given a large number of negatives in the long-form setup, and the fact that MAD is characterized by containing short moments ($4.1$s on average). Although VLG-Net is trained for the task at hand, it achieves comparable or better performance with respect to CLIP only when a strict IoU (IoU${=}0.5$) is considered. However, it lags behind CLIP for most other metrics. We believe the shortcomings of VLG-Net are due to two factors. \emph{(i)} This architecture was developed to ground sentences in short videos, where the entire frame-set can be compared against a sentence in a single forward pass. Thus, it struggles in the long-form setup where we must compare the sentence against all segments of the movie and then aggregate the predictions in a post-processing step. \emph{(ii)} VLG-Net training procedure defines low IoU moments as negatives, thus favoring high performance only for higher IoUs. 

\begin{table}[!t]
    \centering
\setlength{\tabcolsep}{3pt}
\renewcommand{\arraystretch}{1} 
\footnotesize
\begin{tabular}{@{}c@{\hspace{0.4em}} 
l   c@{\hspace{0.2em}} 
cc  c@{\hspace{0.2em}} 
ccc c@{\hspace{0.2em}} 
ccc c@{\hspace{0.2em}} 
c@{}}
\toprule
& 
&\phantom{} & \multicolumn{2}{c}{IoU=0.1} 
& \phantom{} & \multicolumn{2}{c}{IoU=0.3} 
& \phantom{} & \multicolumn{2}{c}{IoU=0.5} 
& \phantom{} \\
\cmidrule{4-5} \cmidrule{7-8} \cmidrule{10-11} 
& Model && R@1 & R@5 
          && R@1 & R@5  
          && R@1 & R@5 \\
\midrule
&Oracle   && $100.00$ &  $-$    
          && $99.88$ &  $-$   
          && $99.42$ &  $-$ \\
          
&Random Chance   && $3.40$  & $15.69$ 
          && $1.47$  &  $7.09$ 
          && $0.52$  &  $2.61$   \\
          
&CLIP~\cite{radford2021learning}  && $20.98$ & $45.49$ 
                                  &&  $9.74$ & $29.63$ 
                                  &&  $4.03$ & $15.90$   \\
                                  
&VLG-Net~\cite{soldan2021vlg}     && $\mathbf{23.94}$ & $\mathbf{51.46}$ 
                                  && $\mathbf{17.51}$ & $\mathbf{43.18}$ 
                                  && $\mathbf{10.17}$ & $\mathbf{30.35}$ \\
                                  
\bottomrule
\end{tabular}

\vspace{-.1cm}
\caption{\label{tab:short_windows}{\bf Short video setup.}
The table showcases the performance of the selected baselines in a short-video setup, where movies are chunked into three minutes (non-overlapping windows). VLG-Net, which falls behind CLIP in the long-form setup, achieves the best grounding performance in most metrics. We can conclude that a new generation of deep learning architectures will have to be investigated to tackle the specific properties of the MAD dataset. 
}
\vspace{-.2cm}

\end{table}

\subsection{The Challenges of Long-form Video Grounding}
This section presents an in-depth analysis of the performance of the selected baselines in the long-form setup. We first investigate how the performance changes when the evaluation is constrained over segments of the movie, whose length is comparable to current datasets. Then explore how methods behave as the size of the movie chunks changes. To this end, we split each video into non-overlapping windows (short videos), and assign the annotations to the short-video with the highest temporal overlap. 

\paragraph{Short-video Setup}
In Table~\ref{tab:short_windows}, we set the short-video window length to three minutes. This duration is a candidate representative for short videos.
The upper bound performance \textit{Oracle} slightly decreases to $99.42\%$ for IoU${=}0.5$. This is a consequence of the division into short videos, which occasionally breaks down a few ground truth moments. The \textit{Random Chance} baseline reports increased performance as the number of proposals generated is reduced. In particular for R@$5$-IoU${=}0.1$, the performance increases from $0.44\%$ (Table~\ref{tab:mad_grounding}) to $15.69\%$ (Table~\ref{tab:short_windows}), demonstrating that the short-video setup is less challenging compared to MAD's original long-form configuration. In a similar trend, performance substantially increases for both CLIP and VLG-Net baselines, with the latter now obtaining the best performances, in all cases.

\begin{figure}[t]
    \centering
        \includegraphics[trim={0cm 0cm 0cm 0cm},width=0.85\linewidth,clip]{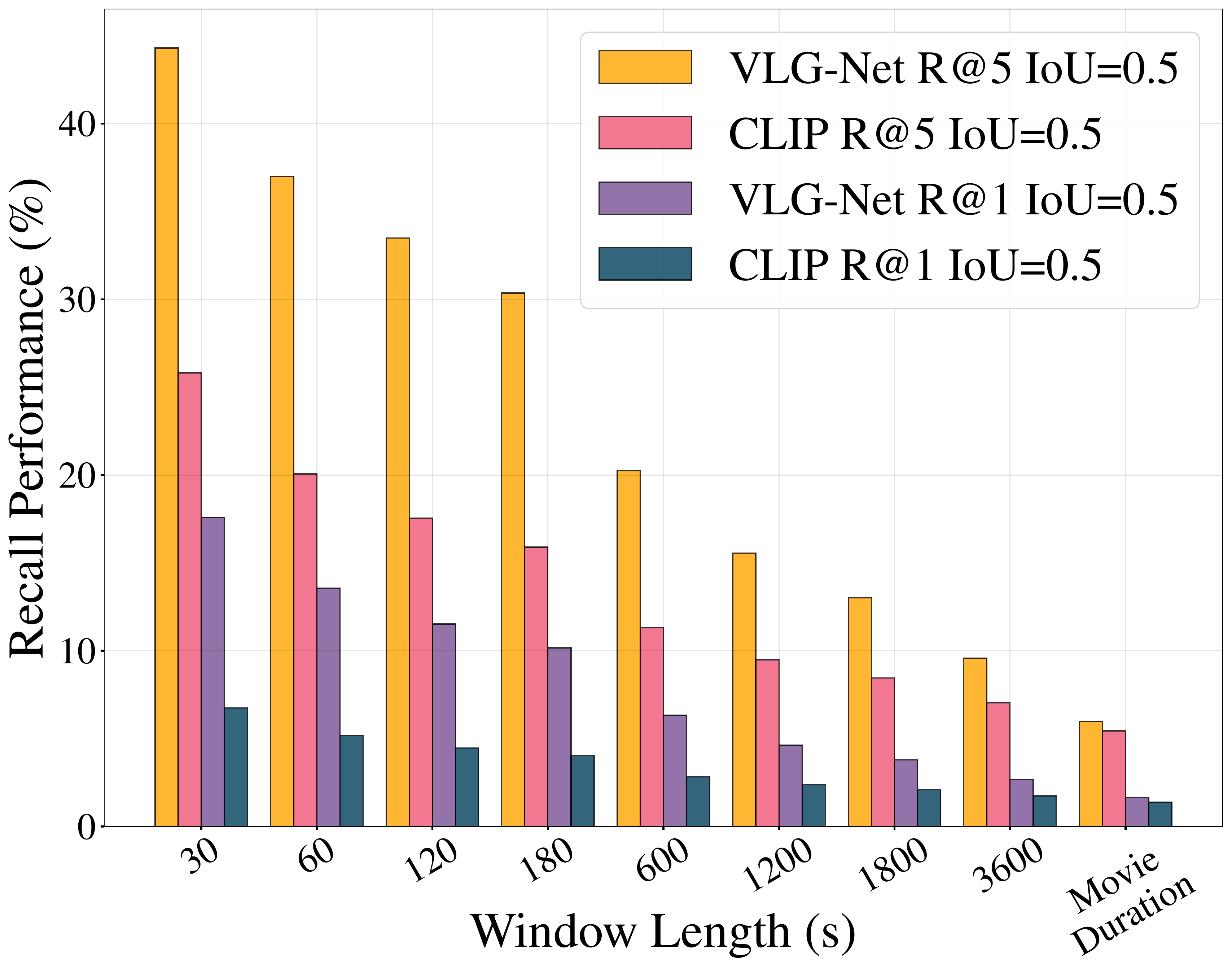}
        
        \vspace{-0.2cm}
        \caption{\textbf{Performance trend across different windows lengths}. 
         We observe from the graph the decrease in performance for both CLIP and VLG-Net, as the evaluation window length increases. This demonstrates that current grounding methods cannot tackle the task in the long-form video setting. 
        \vspace{-0.3cm}
    }
    \label{fig:trend_windows}
\end{figure}

\paragraph{From Short- to Long-form Results}
Figure~\ref{fig:trend_windows} showcases the performance trend for the metrics R@\{$1,5$\}-IoU${=}0.5$, when  the window length is changed from a small value ($30$ seconds) to the entire movie duration (average duration is $2$hrs). The graph displays how the performance steadily drops as the window length increases, showing the challenging setup of long-form grounding enabled by MAD. 

\paragraph{Takeaway}
This set of experiments verifies that VLG-Net could successfully outperform the zero-shot CLIP baseline  when evaluated in a short-video setup.  We can conclude that current state-of-the-art grounding methods are not ready to tackle the long-form setting proposed by MAD. This opens the door to opportunities for the community to leverage previously developed techniques in a more challenging setting and potentially incorporate new constraints when designing deep learning architectures for this task. 

\section{Ablation Study}\label{sec:analysis}
This section focuses on the empirical assessment of the quality of MAD training data.
To help the reader navigate the following experiments, Table~\ref{tab:splits} defines a naming convention associated with different splits of the data. LSMDC16 refers to the original data collected  by~\cite{rohrbach2015dataset,torabi2015using,rohrbach2017movie} for the task of text-to-video retrieval. LSMDC-G is our adaptation of this dataset for the grounding task, as described in Section~\ref{section::test-set}. 
We remind the reader that we could only retrieve $162$ out of $182$ movies included in LSMDC16. 
For these two datasets, we follow the original train/val/test partitions. Finally, we include MAD's split details. 

\begin{table}[!t]
    \centering
\setlength{\tabcolsep}{2pt}
\renewcommand{\arraystretch}{1} 
\resizebox{\linewidth}{!}{%
\footnotesize
\begin{tabular}{lcccc}
    \toprule
    \small
    \multirow{2}{*}{Dataset} & \multirow{2}{*}{Task} & Videos && Annotations \\
    \cmidrule{3-3} \cmidrule{5-5} 
    Name    && Train / Val / Test &&  Train / Val / Test \\
    
    \midrule
    LSMDC16~\cite{rohrbach2017movie} & Retrieval & $155$ / $12$ /~ $17$  && $101.1$ K ~/~~  $7.4$ K ~/~ $10.1$ K \\
    LSMDC-G & Grounding & $138$ / $11$ /~ $13$  && ~~$89.7$  K ~/~~~$6.7$ K ~/~~~  $7.6$ K \\
    MAD     & Grounding & $488$ / $50$ / $112$ && $280.5$ K ~/ $32.1$ K ~/~ $72.0$ K \\
    
    \bottomrule
\end{tabular}
}
\vspace{-.1cm}
\caption{\label{tab:splits}{\bf Data split cheat-sheet.}  
    This table clarifies the data splits used in the following experiments (Table~\ref{tab:ablation_grounding} and Table~\ref{tab:ablation_retrieval}). LSMDC16~\cite{rohrbach2017movie} is the original data collected for retrieval. LSMDC-G is our adaptation to the grounding task. MAD is our proposed dataset. 
}
\vspace{-.5cm}

\end{table}

\begin{table}[!b]
    \vspace{-.2cm}
\centering
\setlength{\tabcolsep}{3pt}
\renewcommand{\arraystretch}{1} 
\resizebox{\linewidth}{!}{%
\footnotesize
\begin{tabular}{cccccccc}
\toprule
\multicolumn{2}{c}{Training Set} && Testing Set && \multicolumn{3}{c}{IoU=0.5} \\

\cmidrule{1-2} \cmidrule{4-4} \cmidrule{6-8} 
\% LSMDC-G & \% MAD && LSMDC-G && R@1 & R@5 & R@10 \\

\midrule
$100\%$ & $0\%$   && Test && $1.36$ & $5.18$  &  $8.82$  \\
$0\%$   & $32\%$  && Test && $0.60$ & $2.60$  &  $5.11$  \\
$0\%$   & $100\%$ && Test && $1.61$ & $6.23$  &  $10.18$  \\
\cmidrule{1-8}
$100\%$ & $32\%$  && Test && $2.18$ & $6.63$  &  $10.73$  \\
$100\%$ & $64\%$  && Test && $2.23$ & $7.79$  &  $11.74$  \\
$100\%$ & $100\%$ && Test && $\mathbf{2.82}$ & $\mathbf{8.74}$  &  $\mathbf{13.36}$  \\

\bottomrule
          
\end{tabular}
}
\vspace{-.1cm}
\caption{\label{tab:ablation_grounding}{\bf Grounding performance with varying training data. }
We investigate VLG-Net~\cite{soldan2021vlg} grounding performance on LSMDC-G test, when different data regimens are used for training. This compares our automatically collected data (MAD training) against the manually curated one (LSMDC-G). 
We conclude that expensive manual curation can be avoided if large scale data is available.
}

\end{table}

\paragraph{Improving Grounding Performance with MAD Data}~\\We are interested in evaluating the contribution our data can bring to the grounding task. We investigate the performance of VLG-Net in the long-form grounding setup when the training data change. All trained models are evaluated on the same test split: LSMDC-G test. 

The first row of Table~\ref{tab:ablation_grounding} shows the performance when VLG-Net is exclusively trained on the LSMDC-G training split. This set only contains data that was manually curated in~\cite{rohrbach2017movie}. The second row is trained with $32\%$ of MAD-training data, which is equivalent to the size of LSMDC-G training split. We observe a drop in performance, which can be associated with the presence of noise introduced by the automatic annotation process in MAD. In the third row, we use the complete MAD training set. Here, MAD's scale allows us to overcome the performance issues associated with noisy data and improve performance to be comparable to only using the clean LSMDC-G for training. Using $100\%$ of MAD data yields a relative improvement of $20\%$ for R@$5$-IoU${=}0.5$. Then, we investigate whether the performance of VLG-Net saturates given the amount of data available. To this end, we use $100\%$ of LSMDC-G training and gradually augment it by adding MAD training samples. In these three experiments (rows 4-6), the performance steadily increases. These results suggest that current models for video grounding do benefit from larger-scale datasets, even though the automatically collected training data may be noisy. Moreover, designing scalable strategies for automatic dataset collection is crucial, as the drawbacks of noisy data can be offset and even overcome with more training data.

\begin{table}[!t]
    \centering
\setlength{\tabcolsep}{3pt}
\renewcommand{\arraystretch}{1} 
\resizebox{\linewidth}{!}{%
\footnotesize
\begin{tabular}{cccccccccc}
\toprule

\multicolumn{2}{c}{Training Set} && Testing Set && \multirow{2}{*}{R@1} & \multirow{2}{*}{R@5} & \multirow{2}{*}{R@10} \\
\cmidrule{1-2} \cmidrule{4-4} 
\% LSMDC16 & \% MAD && LSMDC16 & & & & \\

\midrule
$100\%$ & $0\%$   && Test && $20.9$   &  $39.4$  &  $48.5$  \\
$0\%$   & $36\%$  && Test && $19.2$   &  $35.5$  &  $44.8$  \\
$0\%$   & $100\%$ && Test && $20.5$   &  $38.8$  &  $48.7$  \\
\cmidrule{1-8}
$100\%$ & $36\%$  && Test && $23.3$   &  $40.3$  &  $48.8$  \\
$100\%$ & $72\%$  && Test && $23.6$   &  $\mathbf{41.4}$  &  $49.3$  \\
$100\%$ & $100\%$ && Test &&$\mathbf{24.8}$   &  $40.5$  &  $\mathbf{50.0}$  \\

\bottomrule
          
\end{tabular}
}
\vspace{-.1cm}
\caption{\label{tab:ablation_retrieval}{\bf Retrieval performance on LSMDC16 with model CLIP4Clip~\cite{luo2021clip4clip}.}  
This experiment showcases how MAD data can be valuable for a related task, beyond grounding.  
}
\vspace{-.2cm}

\end{table}

\paragraph{Improving Retrieval Performance with MAD Data}
Adopting the same ablation procedure, we evaluate the possible contribution of our data in a related task, namely text-to-video retrieval. Here, we format our MAD data similarly to LSMDC16, where short videos are trimmed around the annotated timestamps. For this experiment, we use CLIP4Clip~\cite{luo2021clip4clip}, a state-of-the-art architecture for the retrieval task, as the baseline. Table~\ref{tab:ablation_retrieval} reports the performance when different amounts of data is used for training. Again, we see that training with the whole LSMDC16 or MAD leads to a very similar performance. Moreover, our previous argument also holds true in this task. Pouring more data into the task boosts the performance, motivating the benefit of having a scalable dataset like MAD.  

\paragraph{Takeaway}
The MAD dataset is able to boost performance in two closely related tasks, video grounding and text-to-video retrieval, where we show that scale can compensate for potential noise present due to automatic annotation.

\section{Conclusion}\label{sec: conclusions}
The paper presents a new video grounding benchmark called MAD, which builds on high-quality audio descriptions in movies. MAD alleviates the shortcomings of previous grounding datasets. Our automatic annotation pipeline allowed us to collect the largest grounding dataset to date. The experimental section provides baselines for the task solution and highlights the challenging nature of the long-form grounding task introduced by MAD. Our methodology comes with two main hypotheses and limitations:~(i)~Noise cannot be avoided but can be dealt with through scale. (ii)~Due to copyright constraints, MAD's videos will not be publicly released. However, we will provide all necessary features for our experiments' reproducibility and promote future research in this direction.

\noindent\textbf{Acknowledgments.} This work was supported by the King Abdullah University of Science and Technology (KAUST) Office of Sponsored Research through the Visual Computing Center (VCC) funding.

{\small
\bibliographystyle{ieee_fullname}
\bibliography{egbib}
}

\begin{multicols}{2}
\begin{table*}[!b]
    \vspace{-0.3cm}
    \centering
    \resizebox{1.0\textwidth}{!}{
    \begin{tabular}{l||r|r|r||r|r|r|r|r|r|r}
    & \multicolumn{3}{c||}{Videos}    & \multicolumn{7}{c}{Language Queries}       \\ \cmidrule{2-11}
    Split &  Total       & Duration   & Duration   &  Total   & \# Words  &  Total   & \multicolumn{4}{c}{Vocabulary}\\ \cline{8-11} 
                         & Duration  & / Video  & / Moment &  Queries & / Query &  Tokens  & Adj. & Nouns & Verbs & Total  \\ \midrule
\textbf{MAD (Train)}     &  $891.8$ h & $109.65$ min &  $4.0$ s &  $280.5$K  & $13.5$ &  $3.8$M  & $4.8$K & $33.5$K & $12.2$K & $57.6$K  \\
\textbf{MAD (Val/Test)}  &  $315.5$ h & $116.85$ min &  $4.1$ s &  $104.1$K  & $10.6$ &  $1.1$M  & $2.2$K & $11.6$K & $5.8$K & $21.9$K  \\
\bottomrule
    \end{tabular}
    }
    \caption{
    \textbf{Comparison between MAD training and MAD val/test splits.} We verify that the two splits follow similar distributions. We assess that the average video duration, moment length, and sentence length have similar values. Moreover, we highlight how $2/3$ of the video content is reserved for the training split. The size of the training split is also reflected in the total number of queries, with the training set being $2.7\times$ larger than the val/test set. 
    }
    \label{tab:mad_splits_comparison}
    \vspace{-0.3cm}
\end{table*}
\end{multicols}

\begin{multicols}{2}
\begin{figure*}[!b]
    \centering
    \begin{subfigure}[t]{0.335\linewidth}
        \centering
        \includegraphics[width=\linewidth]{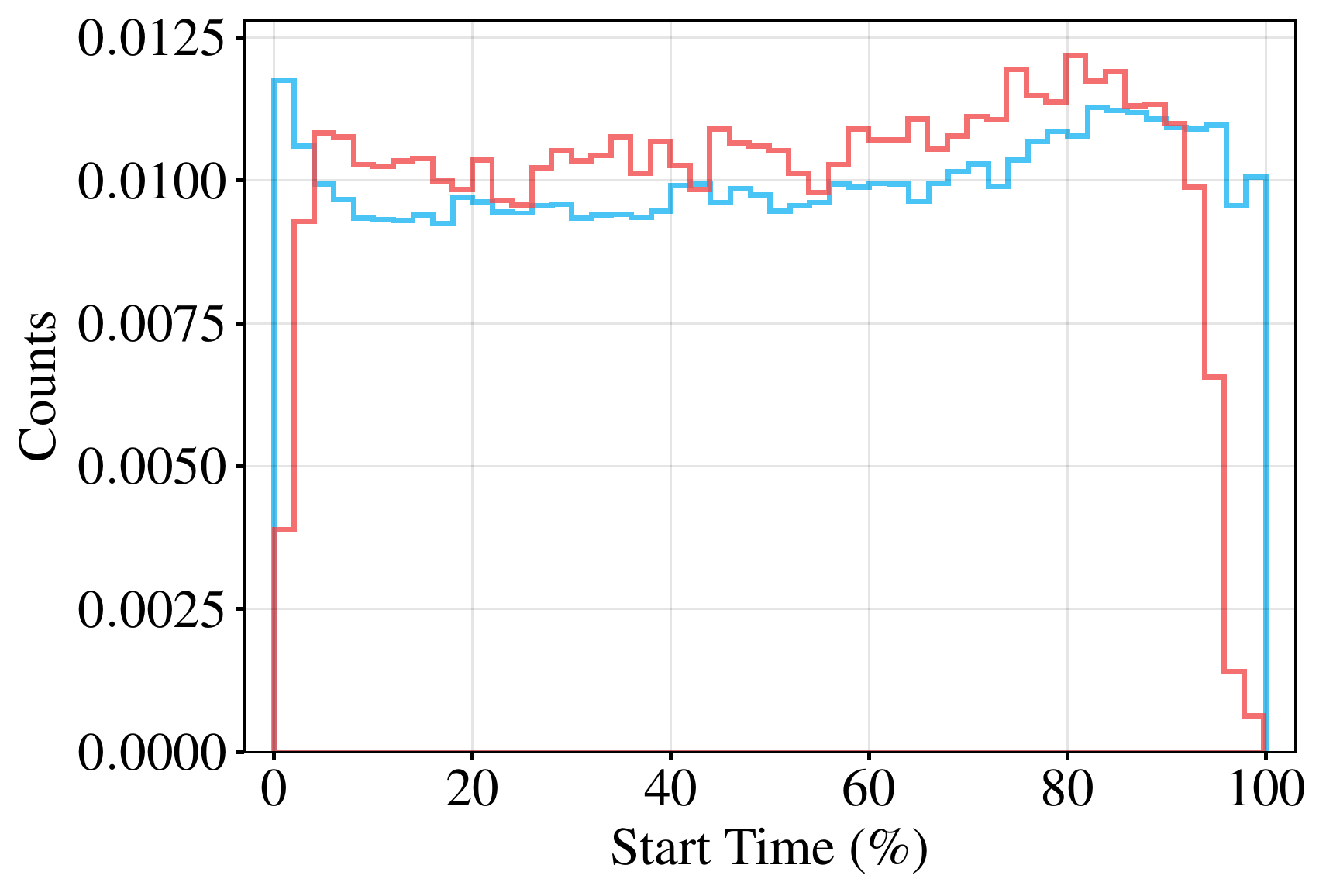}
        \caption{Moment start time histogram}
        \label{fig:starts-sup}
    \end{subfigure}%
    \begin{subfigure}[t]{0.322\linewidth}
        \centering
        \includegraphics[width=\linewidth]{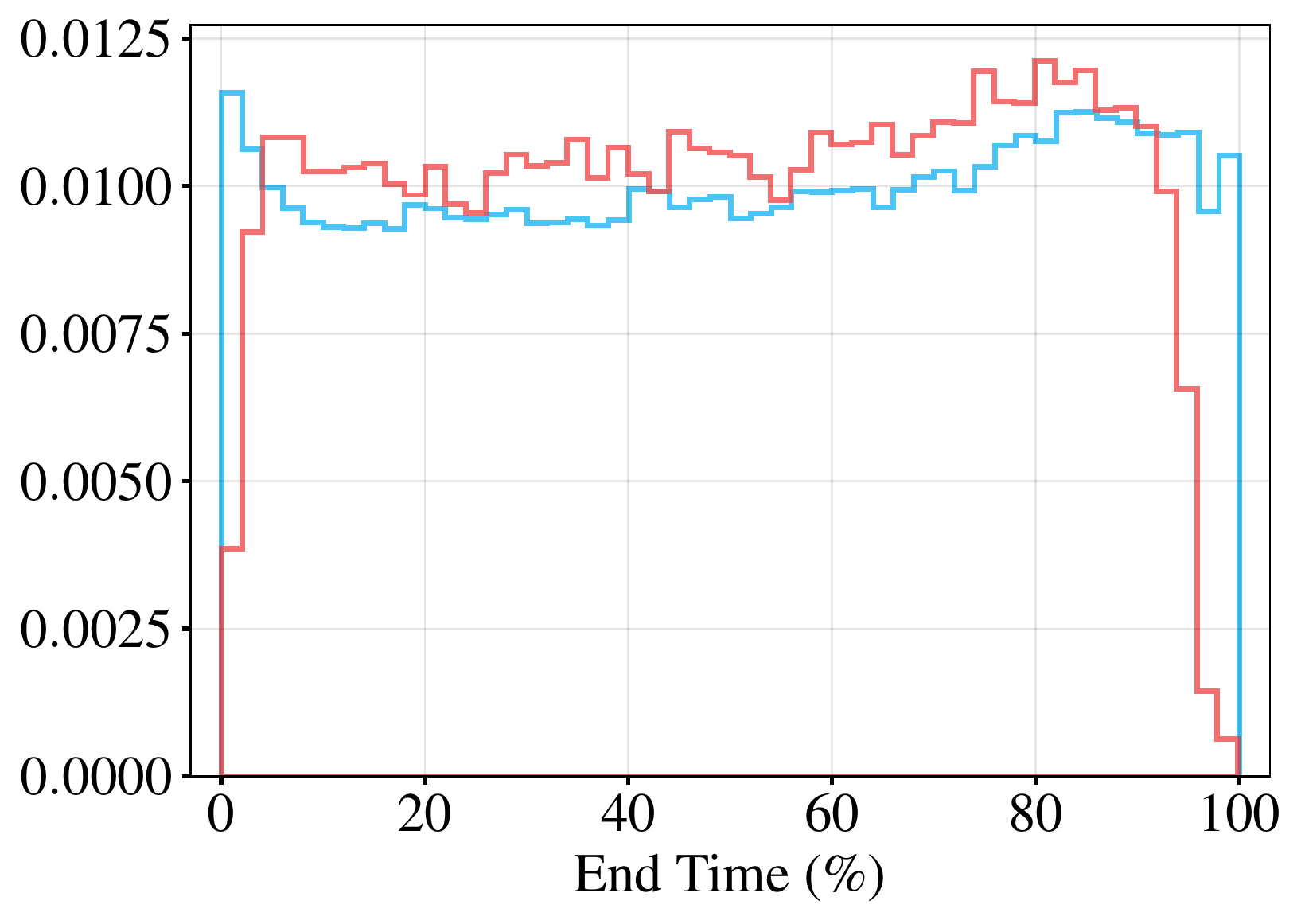}
        \caption{Moment end time histogram}
        \label{fig:ends-sup}
    \end{subfigure}%
    \begin{subfigure}[t]{0.33\linewidth}
        \centering
        \includegraphics[trim=0 0 2.3cm 0.9cm, width=\linewidth]{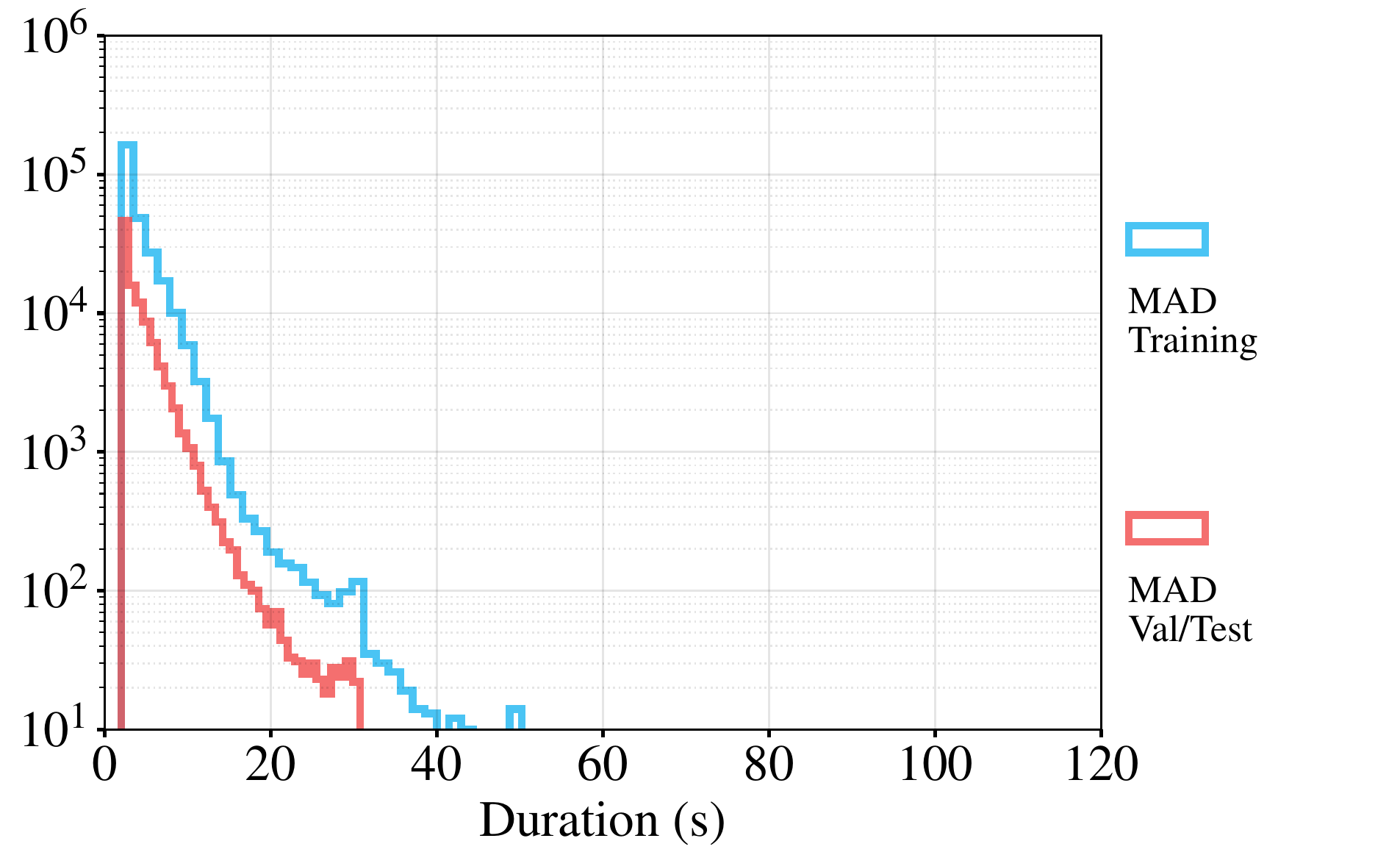}
        \caption{Moment duration histogram}
        \label{fig:durations-sup}
    \end{subfigure}
    \caption{\textbf{Histograms of moment start/end/duration in MAD splits.} The plots represent the normalized (by video length) start/end distributions (a-b), and absolute duration distribution (c) for moments belonging to the training and val/test splits of MAD. The figure showcases that both training and val/test splits follow the same distributions with minor differences between them. 
    }
    \label{fig:dataset-attributes-sup}
\end{figure*}

\end{multicols}

\newpage
\appendix
\section{Appendix}

\subsection{MAD Detailed Statistics}
This section provides additional statistics for the MAD dataset. First, we compare the automatically curated training set against the manually curated validation and test sets, highlighting similarities and differences. Second, we assess the presence of repetitive sentences which might be ambiguous for the language grounding in videos task. We follow by providing additional statistics about MAD's vocabulary and conclude the sections highlighting MAD's large visual and language diversity.

\subsubsection{Data splits comparison}
As described in Section~3 of the main paper, the training set is automatically collected and annotated, whereas the val/test sets of MAD were adapted from the LSMDC dataset~\cite{rohrbach2017movie}. Considering this difference, we analyze the key statistics and discrepancies between the training and the val/test sets in detail. We summarize these results in Table~\ref{tab:mad_splits_comparison} and Figure~\ref{fig:dataset-attributes-sup}.

As shown in Table~\ref{tab:mad_splits_comparison}, the training set contains about $3/4$ of the total video hours and query sentences in MAD, val/test sets contain $1/4$. The average video duration in the two splits is similar, with training videos being, on average, only $6.2\%$ shorter than those in val/test.
Moreover, the average temporal span of the moments is very similar in the two splits, with a difference of only $0.1$ seconds, on average. Regarding the language queries, the training set has slightly longer sentences than the val/test sets, with on average $2.9$ extra words per sentence. We attribute this fact to the automatic annotations procedure of the training set. We observe that sometimes consecutive sentences that are annotated in a short temporal span can be joined together by our annotation pipeline. This does not happen for the val/test set, as sentences were manually refined. \\

Table~\ref{tab:mad_splits_comparison} also highlights a significant difference between the two splits regarding the vocabulary size. The training vocabulary ($57.6$K tokens) is almost three times larger than the one of val/test ($21.9$K tokens). Note that the vocabulary size correlates with the diversity in the language queries. Thus, a more extensive vocabulary is a desirable feature in training, considering that real-world application scenarios might use a variety of words to express similar semantics. 
Finally, the overlap between val/test and training vocabularies is $83\%$, accounting for $18.1$K unique words. 
There are $3.8$K val/test tokens that do not overlap the training set vocabulary. However, these tokens only account for $0.69\%$ of the total tokens in the val/test splits ($1.1M$). 
Moreover, there are $39.5$K unique tokens in the training set that are not present in the val/test. Such unique tokens account for $6.6\%$ of the total training tokens ($3.8M$).  
These features of the dataset will be valuable to evaluate the generalization capabilities of models developed in MAD.

\newpage
Figure~\ref{fig:dataset-attributes-sup} shows the distribution of the relative start time of a moment (Fig.~\ref{fig:starts-sup}) and the relative end time of a moment (Fig.~\ref{fig:ends-sup}). Fig.~\ref{fig:durations-sup} shows the distribution of segments by duration. We show MAD's training split in blue and val/test in red. We observe that the two splits have similar distributions in all three sub-figures. However, we notice that the training set has slightly more moments at the very beginning and at the very end of the videos (Fig.~\ref{fig:starts-sup} and~\ref{fig:ends-sup}). We attribute this discrepancy to the fact that we did not remove the audio descriptions from the movie's opening and credits, as there is not an automatic and reliable way to drop them; LSMDC \textit{manually} removed them. We opt for including such annotations in our dataset. Overall, this design decision has little impact on the data distribution but saves manual effort and keeps our data collection method scalable. For the moment's duration (Fig.~\ref{fig:durations-sup}), both splits exhibit a bias towards short instances and have a long tail distribution, with moments lasting up to $50$ seconds for training and $30$ seconds for val/test.
\begin{figure}[t]
    \vspace{-0.5cm}
    \centering
    \includegraphics[trim={0cm 0cm 0cm 0cm},width=0.85\linewidth,clip]{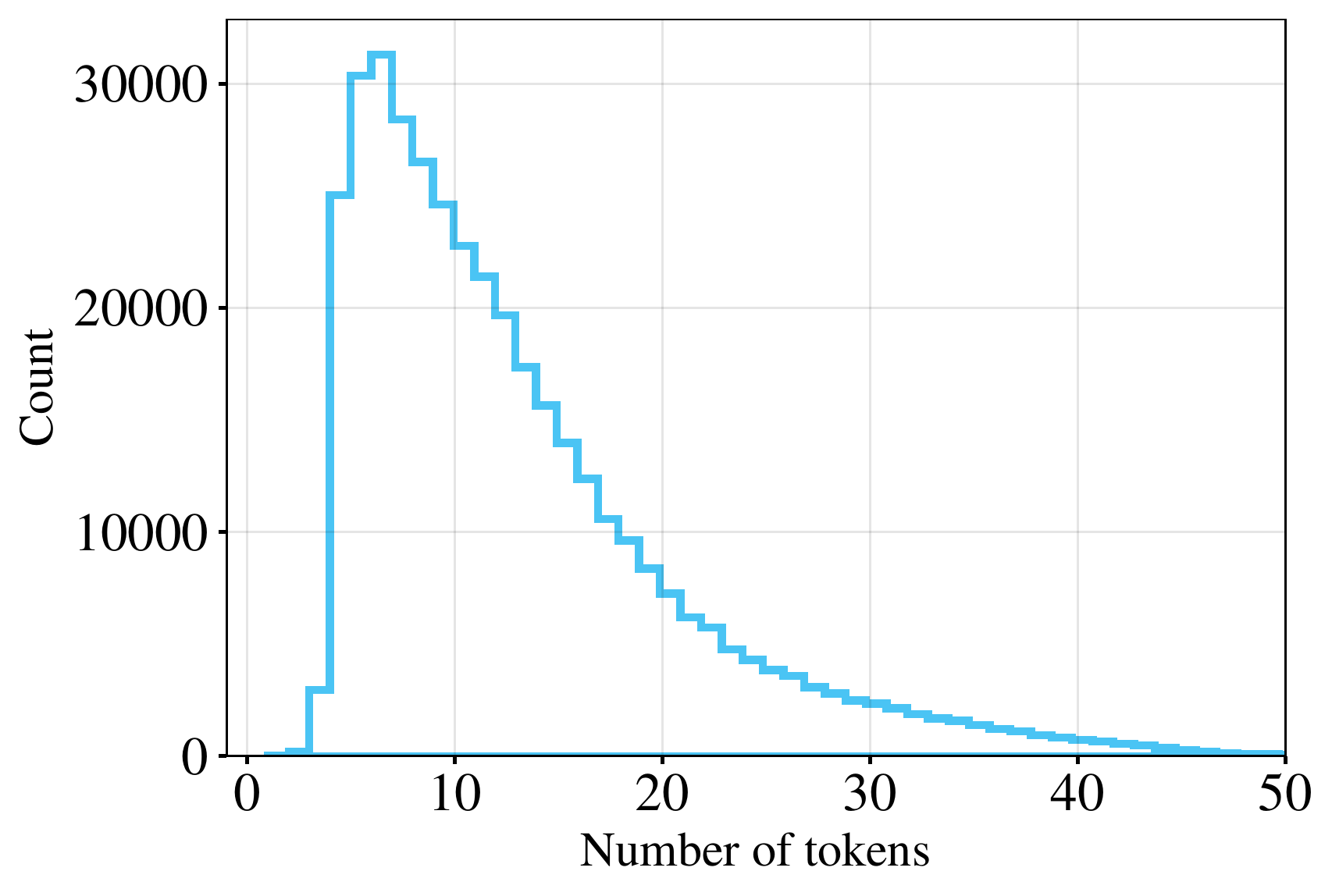}
    \vspace{-0.2cm}
    \caption{\textbf{Sentences length distribution. } Queries length is measured in number of tokens. 
    }
    \label{fig:sentences_length}
    \vspace{-0.5cm}
\end{figure}

\subsubsection{Sentences uniqueness}
Repeating sentences within a movie can be a source of ambiguity for the video-language grounding task. Our automated annotation pipeline, does not enforce individual sentences to be semantically or grammatically different. To quantify this phenomenon, we compute the METEOR similarity score between each pair of sentences within a movie.
The METEOR metric is a widely used metric in NLP~\cite{Krishna_2017_ICCV,rohrbach2014coherent} which correlates well with human judgment on sentence similarity. We use the implementation provided by the NLTK library~\cite{bird2009natural} and empirically observe that the scores are bounded between $[0,1]$. Given these boundaries, we consider a sentence to be unique if its METEOR score with \textbf{every other sentence in the movie} is below $th{=}0.99$.
Following this threshold, $99.7\%$ of sentences can be considered unique. If we lower the threshold to $th{=}0.9$, the uniqueness decreases slightly to $99.2\%$. This suggests that only a few sentences repeat in each movie. 
We emphasize that this estimation cannot directly assess the semantic similarity between sentences, which is a much harder matching problem and requires further research, but remains a good approximation. 

\subsubsection{Additional language statistics}
The MAD dataset contains about $384$K query sentences. The average sentence length is $12.7$ tokens (see Tab 1 in the main paper) with a standard deviation of $8.1$ tokens. 
We show in Figure~\ref{fig:sentences_length} the distribution of the number of tokens per sentence which showcases the variability in query length in the entire dataset.
It is known in the field of computational linguistics that natural language usually follows a long-tailed distribution. We find that it is also the case in the textual annotations of MAD. We compute the frequency distribution of the vocabulary words and find that only $471$ unique tokens out of $61.4$K repeat more than $1000$ times. In comparison, that number increases to $6.3$K if we relax the frequency threshold to only $50$ repetitions. This means that $90\%$ of the tokens in the vocabulary ($55.1$K) appear less than $50$ times in the entire queries corpus. 

\begin{figure}[t]
    \vspace{-0.5cm}
    \centering
    \begin{subfigure}[t]{0.8\linewidth}
        \centering
        \includegraphics[width=\linewidth]{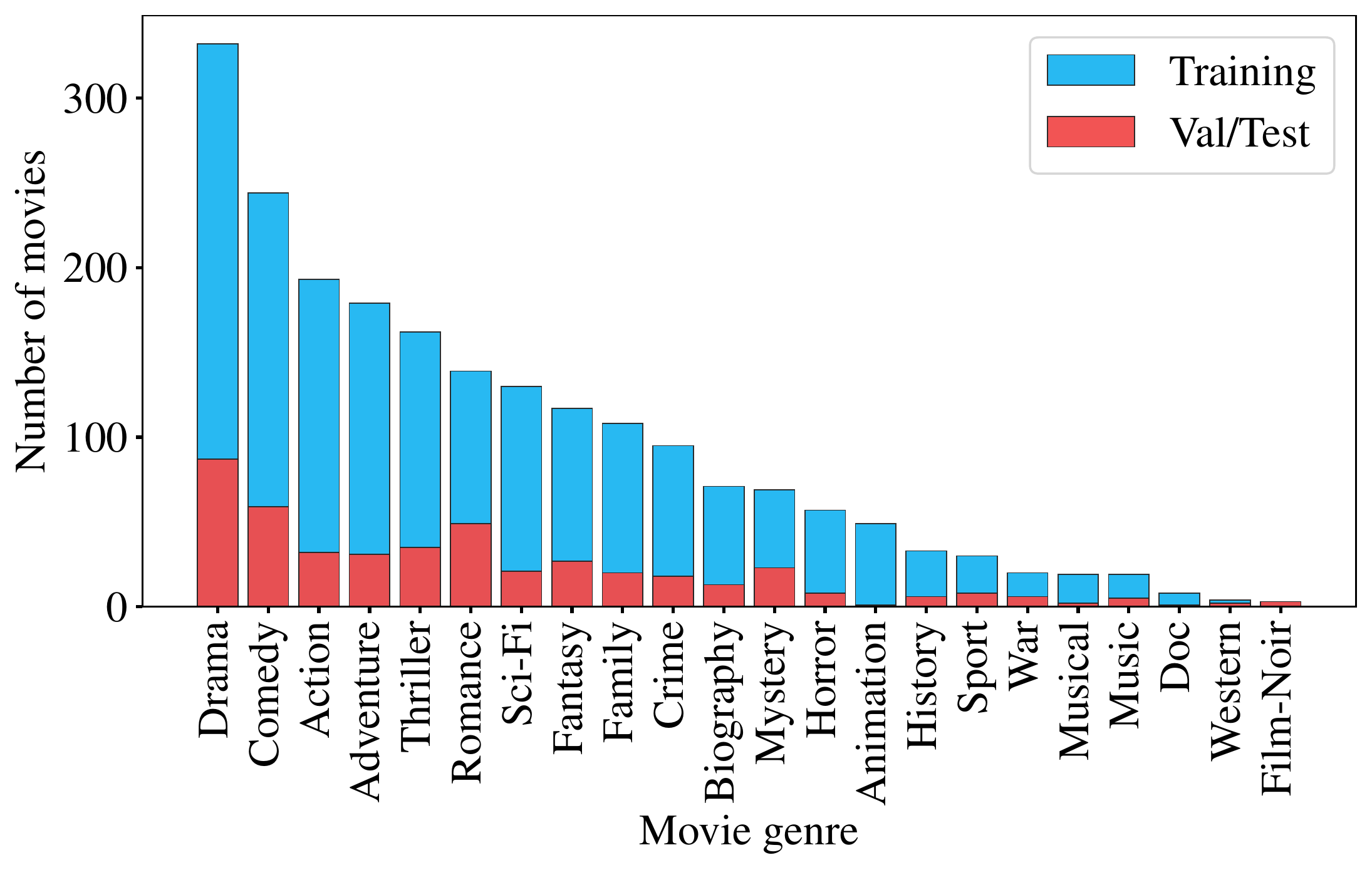}
        \caption{Movies genres.}
        \label{fig:genres}
    \end{subfigure}%
    \\
    \begin{subfigure}[t]{0.8\linewidth}
        \centering
        \includegraphics[width=\linewidth]{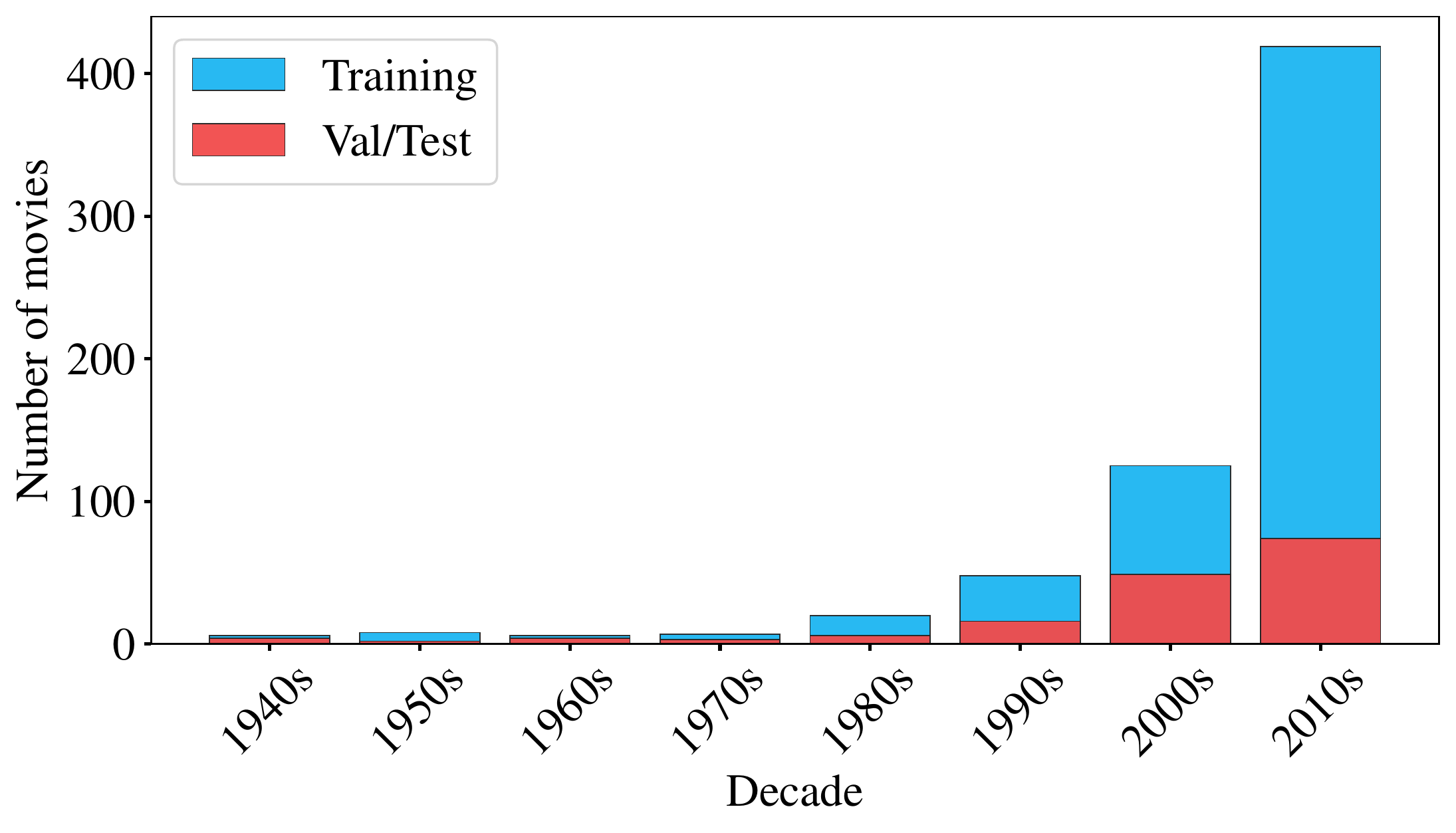}
        \caption{Movies release year.}
        \label{fig:years}
    \end{subfigure}%
    \caption{\textbf{Diversity.} The Figure depicts the wide diversity contained in the dataset. Spanning $22$ different genres and $90$ years of cinema history, MAD presents a highly diverse dataset for the video grounding task. }
    \label{fig:dataset-statistics-supp}
    \vspace{-0.5cm}
\end{figure}

\subsubsection{Diversity}
Figure~\ref{fig:dataset-statistics-supp} shows the distributions of genres and years of MAD movies. We can see that MAD has a wide range in the years the movies are produced (from the 1940s to the last decade) and a large variety of genres. A movie's production year is closely related to its picture quality, filming, edition techniques \cite{pardo2021moviecuts}, character's attire, apparel, action types, etc. The movie genre characterizes how people behave and talk, storytelling techniques, the overall scenes setup, and how fast-paced is the information displayed. These diversities are contained in MAD's videos and descriptions, thus endowing our dataset with a large diversity in video content and related query sentences.


\subsection{VLG-Net Long-Form Adaptation}
In the paper, we select VLG-Net~\cite{soldan2021vlg} as a representative model of the state-of-the-art architectures for the natural language grounding in videos task. 
The challenging long-form nature of the MAD dataset requires some technical changes in the architecture. We detail below the three main upgrades made to this baseline to enable the training and inference over very long videos. 

\textbf{(i) Input.} VLG-Net's default inputs are either frames or snippet-level features that span an entire video. As videos are of different durations, VLG-Net interpolates or extrapolates the features to a predefined length before feeding them to the remaining of the architecture. We change such modeling strategy with the following one: we consider a window of consecutive frames features (\ie, $128$) and input each window independently to the model instead of an entire video. 
Frames are sampled at a constant frame rate (\ie, $5$ frames per second).

During training, for a given sentence and corresponding grounding timestamps, we randomly select a window that contains the annotation's temporal extent. Let us draw an example to understand this approach better. Given a clip's frameset $V=\{v_i\}_{i=1}^{n_v}$ and an associated sentence $S$. We can map the grounding timestamps from the time domain to the frame-index one which we regard as $(t_s,t_e)$ such that $t_s \geq 1$ and $t_e \leq n_v$. At training time, we sample a starting index $(t_s^*)$ in the interval $[t_e - W, t_s]$ and construct our training window as the sequence of frames $\{v_i\}_{i=t_s^*}^{t_s^*+W}$ with $W=128$. Note that $t_e - W \leq ts $.

This process can be seen as a temporal jittering process.
Thanks to this jittering process, the window enclosing the ground-truth segment changes at every epoch (as $t_s^*$ changes at every epoch) for a given sentence. This strategy can be interpreted as a regularization technique that prevents the model from leveraging intra-window biases in the input representation. Moreover, it promotes the model to understand the multi-modal input better and predict the best temporal extent for each language query. 

At inference time, we adopt a sliding window technique, which strides a fixed window over the entire movie. The window size is kept fixed to $128$ frames, and we use a stride of $64$ frames.
For each window, VLG-Net produces a set of proposals with an associated confidence score.
The window-level predictions are then collected and sorted according to the confidence score. The recall metric is measured at the video-level. 

\textbf{(ii) Negative samples.} The original VLG-Net implementation does not make use of negative samples during training. This means that only positive video-language pairs are used. Following the change in the input modeling, the model now only has access to a local portion of the video when making a prediction. Therefore, it is deemed necessary to train the VLG-Net architecture using negatives/unpaired video-language pairs. 
This teaches the model to predict low confidence scores for windows that do not contain visual information relevant to the query being grounded (which are the majority during inference).

Negative samples are defined as a video window ($128$ frames) with IoU is equal to $0$ with the ground truth temporal span of a given sentence. 
With respect to the previous example, a negative video sample is considered as a sequence of consecutive frames of size $W$ which starting index $(t_s^*)$ is sampled outside of the interval $[t_s - W, t_e]$.

At training time, for each sentence, we randomly select a negative sample within the same movie with a probability $p$ or a positive sample (\ie, window containing the ground truth) with probability $1-p$. 
Our experiments show that selecting a negative $70\%$ of the times yields the best performance. 
We do not consider cross-movie negative samples. 
 
\textbf{(iii) Modules.} In Section 4, we described how, to promote a fair comparison against the CLIP baseline~\cite{radford2021learning}, we adopted CLIP's visual and language features as inputs for the VLG-Net baseline. Notably, the language feature extraction strategy poses a technical challenge.
The original sentence tokenizer used by VLG-Net has the capability of extracting syntactic dependencies that are represented as edges in the SyntacGCN module. Because CLIP uses a different tokenizer, we could not retrieve such syntactic dependencies; hence we remove the SyntacGCN module and only retaining the LSTM layers for the language branch.

\end{document}